\documentclass[10pt,twocolumn,letterpaper]{article}

\usepackage{cvpr}              

\usepackage{graphicx}
\usepackage{amsmath}
\usepackage{amssymb}
\usepackage{booktabs}

\usepackage{multirow}
\usepackage{algorithm}
\usepackage{algorithmic}
\usepackage{bm}
\usepackage{amsmath}

\usepackage[pagebackref,breaklinks,colorlinks]{hyperref}

\usepackage[capitalize]{cleveref}
\crefname{section}{Sec.}{Secs.}
\Crefname{section}{Section}{Sections}
\Crefname{table}{Table}{Tables}
\crefname{table}{Tab.}{Tabs.}

\def\cvprPaperID{7558} 
\def\confName{CVPR}
\def\confYear{2022}

\begin{document}

\title{Target-Relevant Knowledge Preservation for \\Multi-Source Domain Adaptive Object Detection}

\author{Jiaxi Wu$^{1,2}$, Jiaxin Chen$^{2}$\footnotemark[1], Mengzhe He$^{3}$, Yiru Wang$^{4}$, Bo Li$^{4}$,\\
Bingqi Ma$^{4}$, Weihao Gan$^{4,5}$, Wei Wu$^{4,5}$, Yali Wang$^{3}$, Di Huang$^{1,2}$\\
$^{1}$State Key Laboratory of Software Development Environment, Beihang University, Beijing, China\\
$^{2}$School of Computer Science and Engineering, Beihang University, Beijing, China\\
$^{3}$Shenzhen Institutes of Advanced Technology, Chinese Academy of Science\\
$^{4}$SenseTime Research \quad\quad $^{5}$Shanghai AI Laboratory\\
{\tt\small \{wujiaxi,jiaxinchen,dhuang\}@buaa.edu.cn, \{hemz, yl.wang\}@siat.ac.cn,}\\
{\tt\small \{libo, mabingqi, wuwei\}@senseauto.com,}
{\tt\small \{wangyiru, ganweihao\}@sensetime.com}
}

\maketitle

\begin{abstract}

Domain adaptive object detection (DAOD) is a promising way to alleviate performance drop of detectors in new scenes. Albeit great effort made in single source domain adaptation, a more generalized task with multiple source domains remains not being well explored, due to knowledge degradation during their combination. To address this issue, we propose a novel approach, namely target-relevant knowledge preservation (TRKP), to unsupervised multi-source DAOD. Specifically, TRKP adopts the teacher-student framework, where the multi-head teacher network is built to extract knowledge from labeled source domains and guide the student network to learn detectors in unlabeled target domain. The teacher network is further equipped with an adversarial multi-source disentanglement (AMSD) module to preserve source domain-specific knowledge and simultaneously perform cross-domain alignment. Besides, a holistic target-relevant mining (HTRM) scheme is developed to re-weight the source images according to the source-target relevance. By this means, the teacher network is enforced to capture target-relevant knowledge, thus benefiting decreasing domain shift when mentoring object detection in the target domain. Extensive experiments are conducted on various widely used benchmarks with new state-of-the-art scores reported, highlighting the effectiveness.
   
\end{abstract}

\section{Introduction}
\label{sec:intro}

\renewcommand{\thefootnote}{\fnsymbol{footnote}}
\footnotetext[1]{Corresponding author.}

In the past decade, convolutional neural networks~\cite{vgg, resnet, resnext} (CNNs) have achieved great progress and delivered significant improvement in visual object detection~\cite{frcnn, ssd, retinanet}. Unfortunately, the well-built detectors suffer from remarkable performance drop when applied to unseen scenes due to domain shift~\cite{ctf, gpa}. Because it is rather expensive and time-consuming to annotate newly collected data, domain adaptive object detection (DAOD)~\cite{wild, ctf, umt} has been receiving increasing attention. It originates from unsupervised domain adaptation (UDA)~\cite{grl, adda, pixelda}, which proves effective in transferring knowledge from the learned domain (known as source domain) to a novel domain (known as target domain) with only unlabeled image for classification. Compared to UDA, DAOD is even more challenging as it  simultaneously locates and classifies all instances of different objects in images with domain shift, requiring generating domain-invariant representations to reduce such a discrepancy in the presence of complex foreground and background variations.


\begin{figure*}[t]
	\centering
	\includegraphics[width=0.95\linewidth]{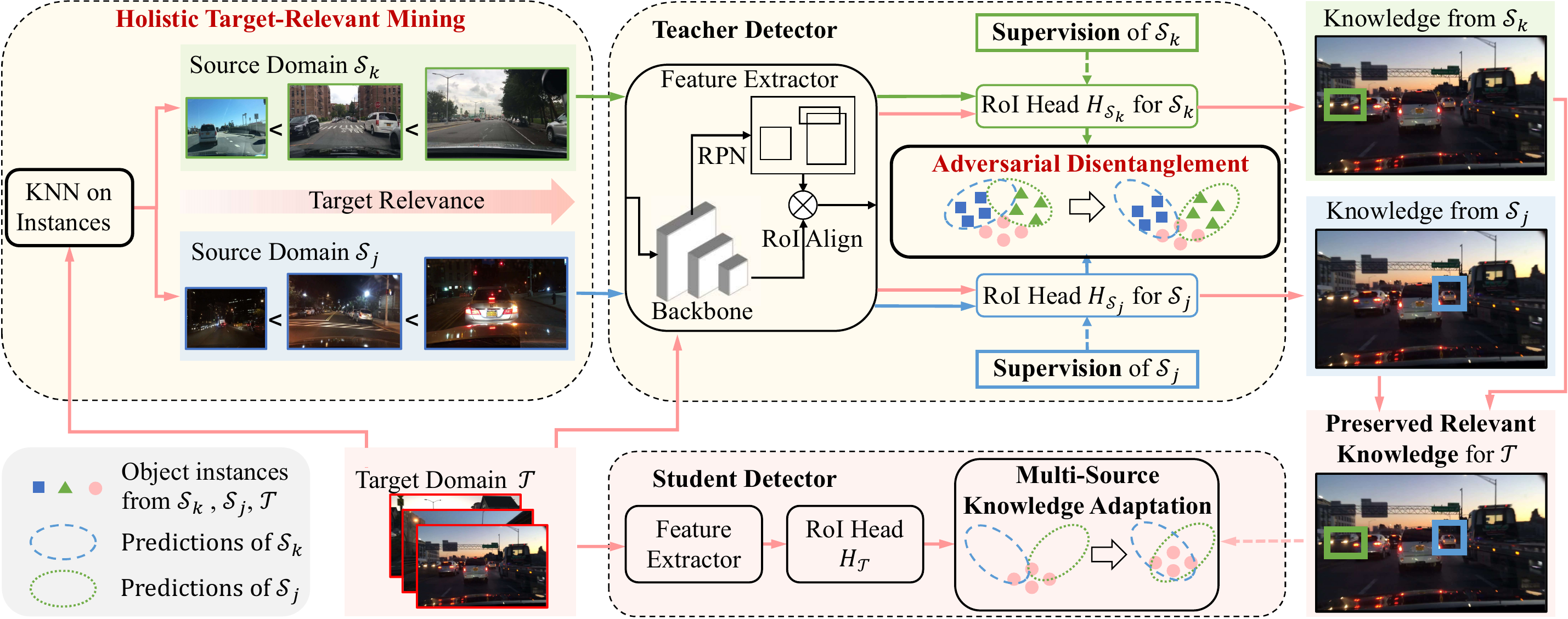}
	\caption{Framework overview of the proposed TRKP approach. The solid arrows refer to forward propagation and the dashed ones denote supervision. The teacher detector is trained on labeled source images and generates pseudo labels for unlabeled images in the target domain, which mentors the student detector. TRKP leverages the adversarial multi-source disentanglement (AMSD) module to preserve source domain-specific knowledge and the holistic target-relevant mining (HTRM) scheme to strengthen encoding target relevance knowledge, which significantly facilitates adapting multi-source knowledge to the target domain.}
	\label{fig:overview}
		\vspace{-0.1cm}
\end{figure*}

Many efforts have been made on DAOD in the literature, and the methods mainly address it in the paradigm of adversarial feature alignment~\cite{sw, diversify, ctf, gpa} or semi-supervised learning~\cite{mtor, cst, umt}. The former directly aligns the features in the source and target domains through adversarial discriminator confused by gradient reversal layer~\cite{sw, ctf}, and it can be fulfilled at the image-level~\cite{wild, diversify}, instance-level~\cite{wild, sw} or/and category-level~\cite{ctf, gpa}. The latter predicts pseudo labels according to the model trained in the source domain and adopts them as guidance to the target domain~\cite{mtor, umt}, and the domain gap can be bridged through enforcing the model consistency. Both the two types of methods show promising results in DAOD for a single pair of source and target.


Multi-source domain adaptation (MSDA) is considered as a more practical scenario in UDA since it assumes that various sources are available for better adaptation to the target domain~\cite{nips18, m3sda, madan}. In addition to the gap between the source and target domains~\cite{nips18,cocktail,mdan}, MSDA also deals with the discrepancy among different sources to avoid negative transfer~\cite{m3sda,secret}. Albeit its prevalence in classification, the multi-source problem has seldom been investigated in detection. To the best of our knowledge, the only attempt is recently given by DMSN~\cite{dmsn}. It follows the pipeline that primarily assigns dynamic weights to multiple sources for alignment and then adapts the compound source to the target in MSDA \cite{distill,m3sda}, and illustrates the necessity of knowledge of different domains to facilitate DAOD. However, there exist two major limitations: (1) the divide-and-merge spindle network conducts early alignment of multiple sources, which often incurs degradation of domain knowledge learned in individual sources for their gaps; (2) the loss memory bank measures target-relevant knowledge in source domains by a temporary discrepancy, leading to a local optimum. Both the facts suggest much room for amelioration.

To tackle the issues aforementioned, this study proposes a novel target-relevant knowledge preservation (TRKP) approach to multi-source DAOD, aiming at enhancing target-relevant knowledge learning from different sources and reducing domain knowledge degradation in adaptation to the target. Specifically, TRKP performs multi-source DAOD in the teacher-student framework, where a multi-head teacher network is constructed to extract knowledge from individual labeled source domains and mentor the student network on detector building in the unlabeled target domain (refer to Fig.~\ref{fig:overview} for an overview). To restrain knowledge degradation, the teacher network embeds an adversarial multi-source disentanglement (AMSD) module to preserve source domain-specific knowledge acquired by corresponding independent detection heads as much as possible during cross-domain alignment. Further, a holistic target-relevant mining (HTRM) scheme is developed to re-weight source images according to source-target relevance. By this means, the teacher network is enforced to capture and highlight target-relevant knowledge at the global level, thus benefiting domain gap decreasing for detector adaptation in the target domain. Extensive experiments are carried out on public benchmarks with state of the art performance reported, demonstrating the advantages of TRKP.

The contributions of this study are three-fold:

1) We propose a novel teacher-student network for multi-source DAOD, which alleviates target-relevant source domain knowledge degradation for alignment through a multi-head teacher structure along with an adversarial source disentanglement module.

2) We propose a target-relevant mining procedure to measure relevance between the source and target domains at the global-level, substantially strengthening target-relevant knowledge acquiring from different sources. 

3) We not only outperform the top counterpart by a large margin in existing protocols, but also achieve a good baseline on a harder scenario with more sources.








\section{Related Work}
\label{sec:related}

\noindent\textbf{Domain Adaptive Object Detection.}
As a well-tuned detector suffers performance degradation when applied to new scenes, unsupervised domain adaptation (UDA) is a promising solution to this dilemma.
Domain adaptive object detection (DAOD) addresses the problem by diminishing the domain shift between seen and unseen scenes~\cite{wild,ctf,epm}.
Most of recent studies can be grouped into two categories: (1) feature alignment based methods that tackle the domain shift by aligning discrepant features in detectors~\cite{wild, sw, gpa, ctf, megacda};
and (2) semi-supervised learning based methods that directly formulate UDA as a semi-supervised learning problem~\cite{mtor, wst-bsr, cst, umt}.
However, these studies are designed on the single-source assumption and fail to deal with multiple source domains.
Here we propose a novel semi-supervised learning based approach specially for multi-source DAOD.

\noindent\textbf{Multi-Source Domain Adaptation.}
The studies on UDA generally focus on alignment between a single pair of source and target domains.
Multi-source domain adaptation (MSDA) considers a more generalized case that multiple source domains are available~\cite{nips18,m3sda,madan}.
It is beneficial to model generalization ability as more diverse data included but more challenging since domain shift also exists among source domains.
There are several early studies~\cite{nips08, nips11, icml13, nips18} handling this problem through a weighted source combination to achieve target-relevant prediction with rigorous theoretical analysis. Recent attempts conduct this re-weighting process in adversarial adaptation~\cite{aggregation, cocktail, mdan}.
Besides, many investigations aim to diminish domain shifts between multiple sources~\cite{m3sda, secret, cvpr21msseg}.
\cite{m3sda} dynamically aligns moments of feature distributions, which consist of pairs of source and target domains and those of source domains.
Rather than explicit feature alignment, \cite{secret} uses pseudo-labeled target samples for implicit alignment.
All the methods above focus on classification, and to the best of our knowledge, DMSN~\cite{dmsn} is the first to introduce MSDA into object detection.
In addition to general DAOD approaches, it develops feature alignment among sources and pseudo subnet learning for their weighted combination.
However, its alignment is limited by knowledge degradation and its temporary domain discrepancy measurement leads to a local optimum.
By contrast, our TRKP aims at preserving more target-relevant knowledge from different source domains to facilitate multi-source DAOD.

\section{Method}
\label{sec:method}
\begin{figure*}[t]
	\centering
	\includegraphics[width=0.95\linewidth]{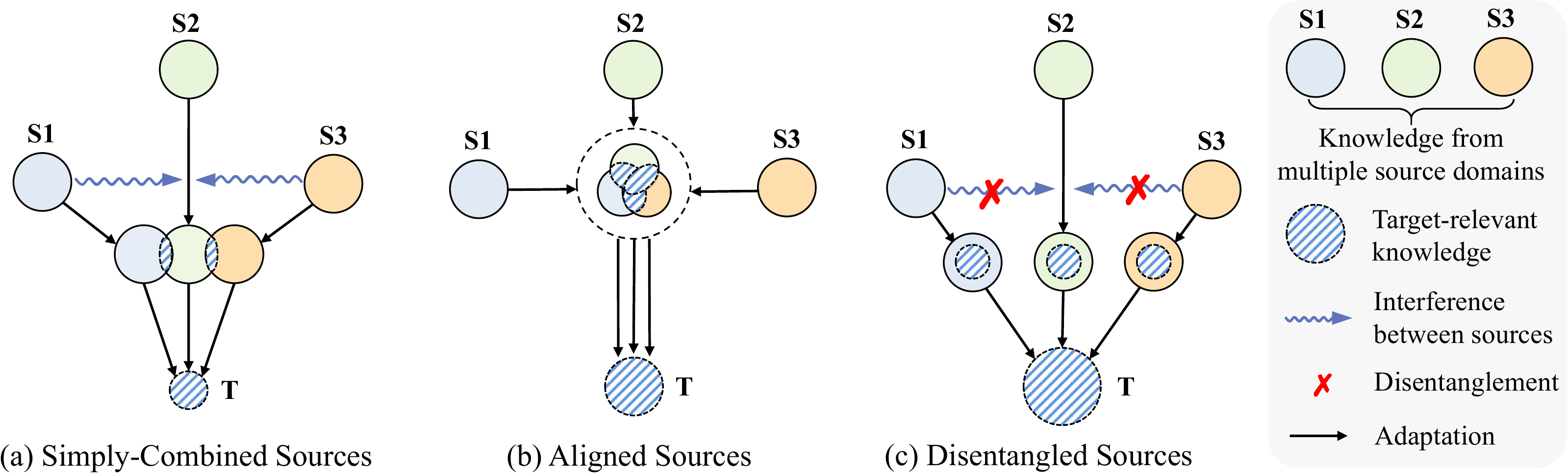}
	\caption{Illustration of different strategies for multi-source adaptation. The area of a circular displays the amount of knowledge. (a) Simply-combined sources probably incurs mutual interference, due to domain shifts among sources. (b) Multi-source alignment reduces the domain shift, but degrades the target-relevant knowledge when performing alignment without the guidance of the target domain. (c) Our method preserves domain-specific target-relevant knowledge by disentangling multiple sources and preventing their mutual interference.}
	\label{fig:adapt}
\end{figure*}

\subsection{Framework Overview}
We firstly describe the problem setting of unsupervised multi-source DAOD and subsequently overview the framework of the proposed approach.

Similar to the general MSDA~\cite{cocktail,m3sda,secret} task, we consider $K$ label-rich source domains $\{\mathcal{S}_{1},\cdots,\mathcal{S}_{K}\}$ and an unlabeled target domain $\mathcal{T}$. Formally, we assume that there exist $N_{\mathcal{S}_{k}}$ labeled images $D_{\mathcal{S}_{k}}=\{(I_{i}^{\mathcal{S}_{k}},~\bm{y}_{i}^{\mathcal{S}_{k}})\}^{N_{\mathcal{S}_{k}}}_{i=1}$ in $\mathcal{S}_{k}$ ($k=1,\cdots,K$), and $N_{\mathcal{T}}$
unlabeled images $D_{\mathcal{T}}=\{I_{i}^{\mathcal{T}}\}_{i=1}^{N_{\mathcal{T}}}$ in $\mathcal{T}$, where $I_{i}^{\mathcal{S}_{k}}$ is the $i$-th image from the $k$-th source domain $\mathcal{S}_{k}$ and $\bm{y}_{i}^{\mathcal{S}_{k}}$ refers to the corresponding label including the bounding boxes and their classes. 


In MSDA, the unsupervised DAOD aims to learn a detector delivering high performance in the unlabeled target domain, by transferring knowledge for detection in $\{\mathcal{S}_{k}\}_{k=1}^{K}$ to $\mathcal{T}$ based on $\{D_{\mathcal{S}_{k}}\}_{k=1}^{K}\mathop{\cup} D_{\mathcal{T}}$. To achieve this goal, we propose a novel approach, namely target-relevant knowledge preservation (TRKP). Inspired by the success of semi-supervised learning in single source DAOD \cite{mtor,umt}, TRKP adopts the teacher-student framework, which proves effective in transferring domain knowledge and bridging the source-to-target gap \cite{mtor,umt}. Specifically, as shown in Fig.~\ref{fig:overview}, TRKP mainly consists of a teacher detector TeDet$(\cdot)$ and a student detector StDet$(\cdot)$, which encodes the knowledge for detection from the source domains and performs object detection in the target domain, respectively. As in \cite{ubt}, StDet$(\cdot)$ adopts the same architecture as TeDet$(\cdot)$. Usually, the `teacher' TeDet$(\cdot)$ is applied to encode knowledge in the source domains by training on $\{D_{\mathcal{S}_{k}}\}^{K}_{k=1}$, and subsequently generate a pseudo label $\bm{\hat{y}}_{j}^{\mathcal{T}}$ for each unlabeled image $I_{j}^{\mathcal{T}}$, which is finally utilized to mentor the `student' StDet$(\cdot)$, \emph{i.e.} training StDet$(\cdot)$ on $\{(I^{\mathcal{T}}_{j},\bm{\hat{y}}_{j}^{\mathcal{T}})\}_{j=1}^{N_{\mathcal{T}}}$. 



As pointed out in~\cite{m3sda,dmsn}, both the multi-source domain shifts and the source-to-target domain gap notably affect the multi-source adaptation to the target domain. DMSN \cite{dmsn} deals with these problems by employing an early multi-source alignment and a local memory bank, which however incurs degradation of knowledge in source domains, thus only reaching a local optimum. To overcome the issues above, we develop an adversarial multi-source disentanglement (AMSD) module together with a holistic target-relevant mining (HTRM) scheme as shown in Fig.~\ref{fig:overview}, which are further incorporated into the teacher-student framework. AMSD enables TeDet$(\cdot)$ to disentangle the single-source knowledge from multiple sources and prevent their mutual interference via adversarial learning, thus fulfilling domain-specific knowledge preservation. HTRM re-weights images from the sources $\{D_{\mathcal{S}_{k}}\}_{k=1}^{K}$ according to their relevance with those from the target $D_{\mathcal{T}}$ in a holistic manner, further facilitating TeDet$(\cdot)$ to encode globally refined target-relevant knowledge. By leveraging both the advantages of AMSD and HTRM, TRKP remarkably alleviates the knowledge degradation, therefore significantly boosting the overall performance. We describe the details of AMSD in Sec.~\ref{sec:disentanglement} and HTRM in Sec.~\ref{sec:mining}, respectively.


\subsection{Adversarial Multi-Source Disentanglement}
\label{sec:disentanglement}
\subsubsection{Knowledge Degradation in MSDA}
\label{sec:problem}

Current approaches for MSDA typically deal with the domain gaps by multi-source combination or alignment. As shown in Fig.~\ref{fig:adapt} (a), the combination based methods bridge the source-target domain gap by taking all the sources as a whole, regardless of their discrepancies. As a consequence, the target-relevant knowledge extracted from one source (\emph{e.g.} S1) may be negatively interfered by another (\emph{e.g.} S2). This kind of knowledge degradation deteriorates the quality of transferred multi-source knowledge. In contrast, as illustrated in Fig.~\ref{fig:adapt} (b), the alignment based approaches pay more attention to removing domain shifts among distinct sources, but probably incur severe loss of knowledge related to the target without the guidance of the target domain, leading to another kind of knowledge degradation.   

As we aim to explore target-relevant knowledge from multiple label-rich sources to train detectors in the unlabeled target domain, both two kinds of knowledge degradation aforementioned should be reduced. There exist several studies emphasizing domain-specific knowledge preservation in heterogeneous domain adaptation~\cite{hda_aaai19,hda_tip,hda_tois} or face recognition under various domain biases~\cite{rl_facedebias,debiasface}, yet not directly applicable to MSDA. This motivates us to present a solution that can jointly preserve domain-specific knowledge and align  the source and target domains as in Fig.~\ref{fig:adapt} (c). We elaborate the details of our solution in Sec.~\ref{subsubsec:KP}.

\subsubsection{Knowledge Preservation via Disentanglement}\label{subsubsec:KP}

\begin{figure}[!t]
	\centering 
	\includegraphics[width=1\linewidth]{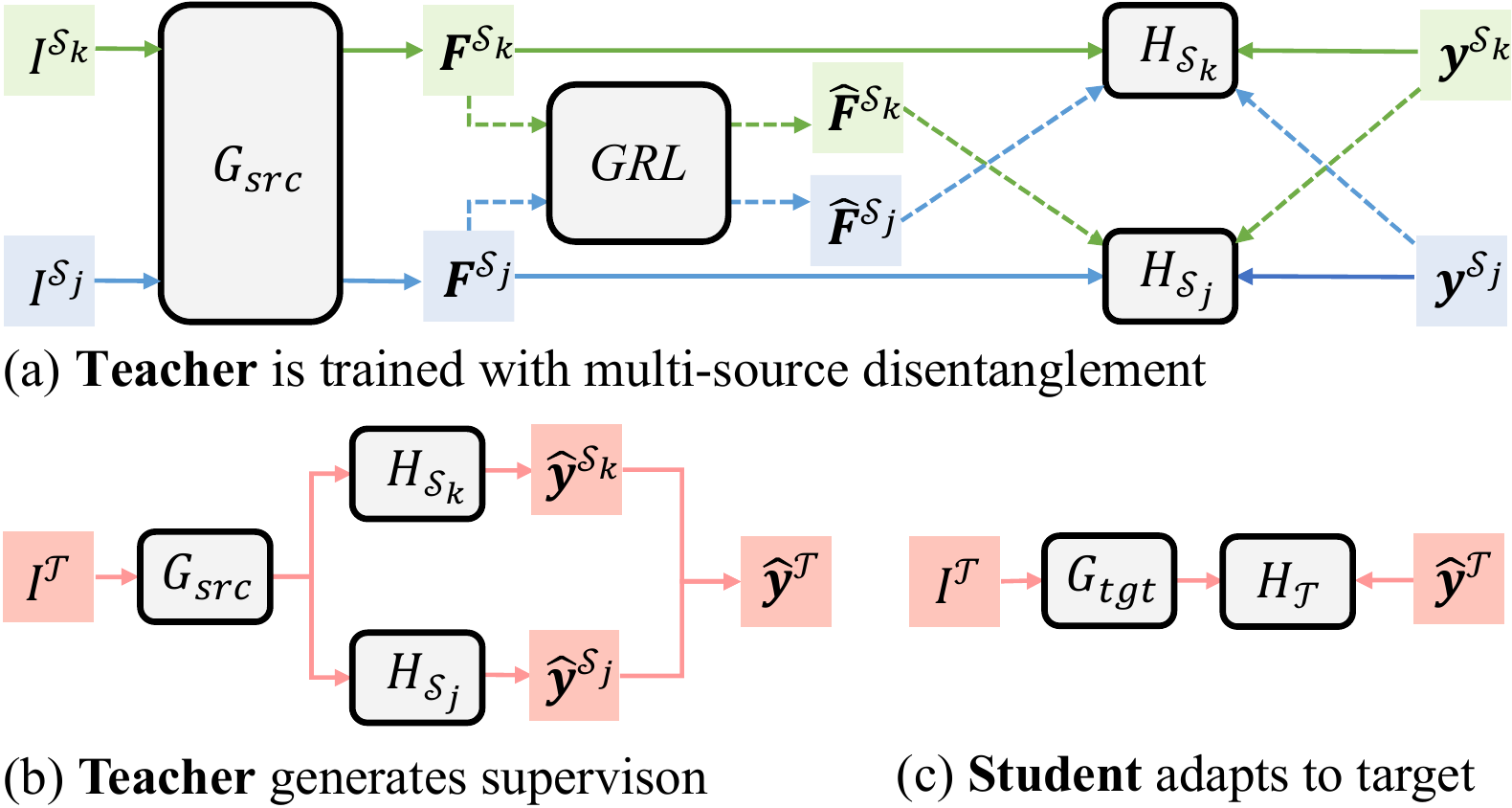}
	\caption{Illustration of the entire training pipeline based on AMSD. The solid arrows refer to teacher-student training and the dashed ones denote disentanglement. (a) The teacher detector is trained on multiple sources with disentanglement. (b) The teacher detector generates pseudo labels for images from the target domain. (c) The student detector adopts pseudo labels for training, and thus accomplishes the multi-source domain adaptation.}
	\label{fig:disentangle}
\end{figure}

In order to alleviate the knowledge degradation, we present AMSD during training TeDet$(\cdot)$ as shown in Fig.~\ref{fig:disentangle}, by encoding the domain-specific knowledge from multiple sources without mutual interference. 

Particularly, we employ the multi-head structure as in \cite{dmsn} in TeDet$(\cdot)$, where each source domain $\mathcal{S}_{k}$ has an individual RoI detection head $H_{\mathcal{S}_{k}}(\cdot)$, but shares the same base network $G_{src}(\cdot)$ (including the backbone and Region Proposal Network known as RPN) with the other source domains. This structure proves effective for its strong generalization ability~\cite{m3sda,secret,dmsn}. Besides, it also facilitates the implementation of multi-source disentanglement and knowledge preservation, since the multiple heads $\{H_{\mathcal{S}_{k}}(\cdot)\}$ have separated parameters for distinct source domains. The student detector StDet$(\cdot)$ adopts the same multi-head architecture as TeDet$(\cdot)$, which is constituted of a base network $G_{tgt}(\cdot)$ and a detection head $H_{\mathcal{T}}(\cdot)$. 

Inspired by~\cite{debiasface}, we disentangle multiple sources by correlation minimization via adversarial learning. Instead of employing additional domain discriminators, we impose constraints on the heads $\{H_{\mathcal{S}_{k}}\}$ and features across source domains, without increasing the model complexity. Specifically, given labeled images $\{(I^{\mathcal{S}_{k}}_{i},\bm{y}^{\mathcal{S}_{k}}_{i})\}$ from multiple sources, the corresponding deep features are fetched by $G_{src}$, denoted as $\{\bm{F}^{\mathcal{S}_{k}}_{i}={G_{src}}(I^{\mathcal{S}_{k}}_{i})\}$. A gradient reverse layer $GRL(\cdot)$ is introduced between the feature extractor $G_{src}$ and heads $\{H_{\mathcal{S}_{k}}\}$ to implement adversarial learning. In the forward propagation of $GRL$, an adversarial feature $\bm{\hat{F}}^{\mathcal{S}_{k}}_{i}=GRL(\bm{F}^{\mathcal{S}_{k}}_{i})$ is generated for an input $\bm{F}^{\mathcal{S}_{k}}_{i}$. In the back propagation of $GRL$, the sign of the input gradient is simply reversed and multiplied by a factor $\mu$. To facilitate learning domain-specific knowledge from the $k$-th source domain $\mathcal{S}_{k}$, we formulate the following loss w.r.t. the $k$-th detection head $H_{\mathcal{S}_{k}}$: 
\begin{equation} 
\label{eq:1}
\mathcal{L}^{H_{\mathcal{S}_{k}}}_{i}=l[H_{\mathcal{S}_{k}}(\bm{F}^{\mathcal{S}_{k}}_{i})] +\frac{\lambda}{K-1}\sum_{j\neq{k}}^{K}{l[H_{\mathcal{S}_{j}}(\bm{\hat{F}}^{\mathcal{S}_{k}}_{i})]}, 
\end{equation}
where $l[\cdot]$ is the conventional detection loss (\eg, the focal loss and smooth $L_{1}$ loss), and $\lambda$ is a trade-off parameter. The label $\bm{y}^{\mathcal{S}_{k}}_{i}$ is simply omitted here for succinctness. 

As observed from Eq.~\eqref{eq:1}, the standard detection loss $l[H_{\mathcal{S}_{k}}(\bm{F}^{\mathcal{S}_{k}}_{i})]$ trains $H_{\mathcal{S}_{k}}$ by using the feature from $\mathcal{S}_{k}$, thus encoding knowledge from $\mathcal{S}_{k}$. The additional loss $l[H_{\mathcal{S}_{j}}(\bm{\hat{F}}^{\mathcal{S}_{k}}_{i})]$ measures the discrepancy between the ground-truth label and the prediction by the head $H_{\mathcal{S}_{j}}$ using the adversarial feature $\bm{\hat{F}}^{\mathcal{S}_{k}}_{i}$ from a distinct source domain $\mathcal{S}_{j}$ ($j\neq i$). Recall that the gradient w.r.t. $\bm{\hat{F}}^{\mathcal{S}_{k}}_{i}$ is reversed via $GRL$ in back propagation. Therefore, minimizing $l[H_{\mathcal{S}_{j}}(\bm{\hat{F}}^{\mathcal{S}_{k}}_{i})]$ will increase the prediction error made by $H_{\mathcal{S}_{j}}$ on $\bm{F}^{\mathcal{S}_{k}}_{i}$. In other words, the loss $\mathcal{L}^{H_{\mathcal{S}_{k}}}_{i}$ in Eq.~\eqref{eq:1} enforces $H_{\mathcal{S}_{k}}$ to encode domain-specific knowledge from $\mathcal{S}_{k}$ and simultaneously puzzles the other heads $H_{\mathcal{S}_{j}}$ ($j\neq i$) by forcing them to yield distinct predictions. 

Based on Eq.~\eqref{eq:1}, the teacher detector is trained as below:
\begin{equation}
\label{eq:2}
\min_{G_{src},\{H_{\mathcal{S}_{k}}\}_{k=1}^{K}}\sum_{k=1}^{K}\sum_{i=1}^{N_{\mathcal{S}_{k}}}\mathcal{L}_{i}^{H_{\mathcal{S}_{k}}}.
\end{equation}

As being optimized in \cref{eq:2}, each head $H_{S_{k}}$ is disentangled from the other sources, thus encoding domain-specific knowledge. By this means, the mutual interference between sources can be mitigated, benefiting decreasing knowledge degradation. 

\subsubsection{Multi-Source Knowledge Adaptation}

After training the teacher detector TeDet$(\cdot)$ by AMSD, the domain-specific knowledge encoded in each head is subsequently adapted to the target domain via training the student detector StDet$(\cdot)$. Concretely, given an unlabeled image $I_{j}^{\mathcal{T}}$ from the target domain, each head $H_{\mathcal{S}_{k}}$ separately generates a prediction $\bm{\hat{y}}_{j}^{\mathcal{T},\mathcal{S}_{k}}$, and the averaged one (conducted on RoI) $\bm{\hat{y}}_{j}^{\mathcal{T}}=\frac{1}{K}\sum_{k}^{K}\bm{\hat{y}}_{j}^{\mathcal{T},\mathcal{S}_{k}}$ is utilized as the pseudo label. Finally, the `student' StDet$(\cdot)$ is mentored by TeDet$(\cdot)$ via the following optimization process based on $\{(I^{\mathcal{T}}_{j},\bm{\hat{y}}_{j}^{\mathcal{T}})\}_{j=1}^{N_{\mathcal{T}}}$:
\begin{equation}
\label{eq:3}
\min_{G_{tgt},H_{\mathcal{T}}}\sum_{j=1}^{N_{\mathcal{T}}}l[H_{\mathcal{T}}(G_{tgt}(I^{\mathcal{T}}_{j}))].
\end{equation}

During training StDet$(\cdot)$ based on \cref{eq:3}, the multi-source domains and the target domain are implicitly aligned. However, training the `student' StDet$(\cdot)$ with a fixed `teacher' TeDet$(\cdot)$ tends to incur overfitting~\cite{meanteacher}. The Exponential Moving Average (EMA) \cite{ubt} mechanism addresses this issue by regularizing the learning of TeDet$(\cdot)$ with the gradient of StDet$(\cdot)$. We therefore employ it in our framework to fulfill the multi-source knowledge adaptation in a more effective way.

\subsection{Holistic Target-Relevant Mining}
\label{sec:mining}

As observed in Eq.~\eqref{eq:2}, images from multiple sources are treated equally when training TeDet$(\cdot)$. Due to the lack of guidance of the target, images that are less relevant to the target domain are given the same importance as more relevant ones, which deteriorates the quality of knowledge adaption. Previous works in MSDA~\cite{nips11,icml13,nips18} tackle this problem by using a distribution-weighted combination specially designed for classification, which is not fully suitable for object detection. DMSN~\cite{dmsn} makes the first attempt in detection by proposing a dynamic loss memory bank to measure the discrepancy between the source and target domains. Nevertheless, it only captures local relevance information in mini-batches, leading to a local optimal solution.

To address the issue above, we develop HTRM to guarantee that the teacher detector encodes target-relevant knowledge at the global level, by assigning each source image $I_{i}^{\mathcal{S}_{k}}$ a target-relevant weight $\alpha_{i}^{\mathcal{S}_{k}}$. To achieve this goal, we first extract the deep feature $\bm{F}_{i}^{\mathcal{S}_{k}}$ via $G_{src}(\cdot)$ for each image $I_{i}^{\mathcal{S}_{k}}$. To avoid the interference from massive backgrounds, we only select the RoI features locating in the object area according to the label $\bm{y}_{i}^{\mathcal{S}_{k}}$, which are further pooled as a set of features denoted by $\{\bm{f}_{i,j}^{\mathcal{S}_{k}}\}_{j=1}^{|\bm{y}_{i}^{\mathcal{S}_{k}}|}$. Here, $|\bm{y}_{i}^{\mathcal{S}_{k}}|$ stands for the number of annotated bounding boxes in the $i$-th image $I_{i}^{\mathcal{S}{k}}$. By repeating this procedure, we finally obtain the instance-level feature set for all the images from the multi-source domains, denoted by $\mathcal{G}=\{\{\{\bm{f}_{i,j}^{\mathcal{S}_{k}}\}_{j=1}^{|\bm{y}_{i}^{\mathcal{S}_{k}}|}\}_{i=1}^{N_{\mathcal{S}_{k}}}\}_{k=1}^{K}$ . Similarly, based on the pseudo labels $\{\bm{\hat{y}}_{m}^{\mathcal{T}}\}$ and $G_{tgt}(\cdot)$ of the student detector, we extract the instance-level feature set from the target domain, denoted by $\mathcal{Q}=\{\{\bm{f}_{n,m}^{\mathcal{T}}\}_{m=1}^{|\bm{\hat{y}}_{n}^{\mathcal{T}}|}\}_{n=1}^{N_{\mathcal{T}}}$.

We follow~\cite{icml13} by applying the nearest neighbor algorithm to mine cross-domain relevance $\{\alpha_{i}^{\mathcal{S}_{k}}\}$. As summarized in Algorithm \ref{alg:1}, the mining process mainly consists of two steps: 1) for each feature $\bm{f}_{n,m}^{\mathcal{T}}\in \mathcal{Q}$ from the target domain, we search its $K'$ nearest neighbors $\mathcal{N}_{\bm{f}^{\mathcal{T}}_{n,m}}$ in $\mathcal{G}$ from the source domains, where the cosine distance is used as the similarity metric; 2) for the $i$-th image $I_{i}^{\mathcal{S}_{k}}$ from the $k$-th source domain represented by $\{\bm{f}_{i,j}^{\mathcal{S}_{k}}\}_{j=1}^{|\bm{y}_{i}^{\mathcal{S}_{k}}|}$, we compute the frequency $w_{i}^{\mathcal{S}_{k}}$ by counting the number of elements in $\mathcal{Q}$ that include at least one member in  $\{\bm{f}_{i,j}^{\mathcal{S}_{k}}\}_{j=1}^{|\bm{y}_{i}^{\mathcal{S}_{k}}|}$ as $K'$ nearest neighbors. Note that $w_{i}^{\mathcal{S}_{k}}$ in step 2) is computed by using the holistic feature set from the target domain, thus mining the target-relevance in a global view. Based on $w_{i}^{\mathcal{S}_{k}}$, the relevance weight $\alpha_{i}^{\mathcal{S}_{k}}$ is formulated as below:

\begin{algorithm}[!t]
	\caption{Holistic Target-Relevant Mining}
	\label{alg:1}
	\textbf{Input}: The object-level feature set $\mathcal{G}$ from multiple source domains and the feature set $\mathcal{Q}$ from the target domain; the hyper-parameter $K'$.\\
	\textbf{Output}: The relevance weights $\{\alpha_{i}^{\mathcal{S}_{k}}\}$ of the source images w.r.t the target domain.\\
	\textbf{Initialize}: $w_{i}^{S_{k}}:=0.$
	\begin{algorithmic}[1] 
		\FOR{$\bm{f}^{\mathcal{T}}$ in $\mathcal{Q}$}
		\STATE Find the $K'$-nearest neighbors of $\bm{f}^{\mathcal{T}}$ in $\mathcal{G}$ as $\mathcal{N}_{\bm{f}^{\mathcal{T}}}$
		\FOR{$\bm{f}_{i,j}^{\mathcal{S}_{k}}$ in the neighborhood $\mathcal{N}_{\bm{f}^{\mathcal{T}}}$}
		\STATE $w_{i}^{\mathcal{S}_{k}}:=w_{i}^{\mathcal{S}_{k}}+1$
		\ENDFOR
		\ENDFOR
		\\
		\STATE Compute the weight $\{\alpha_{i}^{\mathcal{S}_{k}}\}$ based on $\{w_{i}^{\mathcal{S}_{k}}\}$ and Eq.~\eqref{eq:5}
	\end{algorithmic}
\end{algorithm}

\begin{equation}
\label{eq:5}
\alpha_{i}^{\mathcal{S}_{k}}=
\begin{cases}
\gamma\log(\frac{w_{i}^{\mathcal{S}_{k}}}{K'})+\beta,\quad &w_{i}^{\mathcal{S}_{k}}>{K'},\\
0,\quad &w_{i}^{\mathcal{S}_{k}}\leq{K'},
\end{cases}
\end{equation}
where $\gamma$ and $\beta$ control the magnitude of $\alpha_{i}^{\mathcal{S}_{k}}$. From Eq.~\eqref{eq:5}, we can observe that $\alpha_{i}^{\mathcal{S}_{k}}$ becomes large if the source image $I_{i}^{\mathcal{S}_{k}}$ is closely relevant to the target, and turns to $0$ otherwise. 

Based on $\{\alpha_{i}^{\mathcal{S}_{k}}\}$, we can re-weight the importance of images from multiple sources as illustrated in Fig.~\ref{fig:overview}, and apply it to train a target-relevant teacher detector by reformulating the loss function in Eq.~\eqref{eq:2} as the following:
\begin{equation}
\label{eq:6}
\min_{G_{src},\{H_{\mathcal{S}_{k}}\}_{k=1}^{K}} \sum_{k=1}^{K}\sum_{i=1}^{N_{\mathcal{S}_{i}}}\alpha_{i}^{\mathcal{S}_{k}}\mathcal{L}_{i}^{H_{\mathcal{S}_{k}}}.
\end{equation}

Based on Eq.~\eqref{eq:6}, TeDet$(\cdot)$ is explicitly enforced to learn from target-relevant samples, and thus restrains from the interference from the information irrelevant to the target.


\section{Experiments}
\label{sec:exp}

In this section, we evaluate the performance of TRKP by following the settings in \cite{dmsn}, including the cross camera adaptation in Sec.~\ref{sec:exp1} and the cross time adaptation in Sec.~\ref{sec:exp2}.
In addition, we present a new setting, which contains more sources with mixed domain gaps in Sec.~\ref{sec:exp3}. We also conduct ablation studies as summarized in Sec.~\ref{sec:exp4}

\noindent\textbf{Implementation Details.}
Similar to~\cite{ctf,dmsn}, we adopt Faster R-CNN~\cite{frcnn} with RoI Align~\cite{maskrcnn} and VGG16~\cite{vgg} backbone as the basic detector to make fair comparisons.
All the input images are resized such that the shorter lengths have 600 pixels. As for the teacher-student learning framework, we adopt the same settings as in UBT~\cite{ubt}, which is a representative of semi-supervised object detection. Concretely, the confidence threshold for pseudo labeling is set to 0.7. The smoothing coefficient in EMA is set as 0.9999. For AMSD, the hyper-parameters $\lambda$ and $\mu$ are fixed to 0.2 and 0.01, respectively. For HTRM, the number of nearest neighbors $K'$ is set to 5. The scaling factors $\gamma$ and $\beta$ in Eq.~\eqref{eq:5} are fixed as 1.0 and 0.5 by default. The learning rate is 0.01 with the batch size at 16. We utilize 20 epochs in training, where the teacher detector is trained individually for the first 10 epochs, after which HTRM is conducted to re-weight source images, followed by training StDet$(\cdot)$ for domain adaptation. All the experiments are carried out on 8 NVIDIA 1080Ti GPUs.

\noindent\textbf{Comparative Approaches.}
We compare TRKP to the following state-of-the-art approaches: (1) Source-only method which applies the basic Faster R-CNN~\cite{frcnn} detector without adaptation to the target domain; (2) Single-Source \& Source-Combined methods including SW~\cite{sw}, GPA~\cite{gpa}, UMT~\cite{umt} and UBT~\cite{ubt}, which conduct DAOD with the single-source assumption; (3) MSDA methods including MDAN~\cite{mdan}, M$^{3}$SDA and DMSN~\cite{dmsn}. We also report the performance of Oracle trained by fully labeled target images, as an estimated upper bound.

\subsection{Cross Camera Adaptation}
\label{sec:exp1}

\begin{table}[!t]
	\centering
	\footnotesize
	\begin{tabular}{c|ccc|c|c}
		\hline
		Setting & \multicolumn{3}{c|}{Source} & Method & AP \\ \hline
		\multirow{3}{*}{Source Only} & \multicolumn{3}{c|}{C} & \multirow{3}{*}{FRCNN~\cite{frcnn}} & 44.6 \\
		& \multicolumn{3}{c|}{K} &  & 28.6 \\
		& \multicolumn{3}{c|}{C+K} &  & 43.2 \\ \hline
		\multirow{4}{*}{Single Source} & \multicolumn{3}{c|}{\multirow{4}{*}{C}} & SW~\cite{sw} & 45.5 \\
		& \multicolumn{3}{c|}{} & CRDA~\cite{crda} & 46.5 \\
		& \multicolumn{3}{c|}{} & UMT~\cite{umt} & 47.5 \\
		& \multicolumn{3}{c|}{} & UBT~\cite{ubt} (Baseline)& 48.4 \\ \hline
		\multirow{4}{*}{Single Source} & \multicolumn{3}{c|}{\multirow{4}{*}{K}} & SW~\cite{sw} & 29.6 \\
		& \multicolumn{3}{c|}{} & CRDA~\cite{crda} & 30.8 \\
		& \multicolumn{3}{c|}{} & UMT~\cite{umt} & 35.4 \\
		& \multicolumn{3}{c|}{} & UBT~\cite{ubt} (Baseline)& 33.8 \\ \hline
		\multirow{4}{*}{Source Combined} & \multicolumn{3}{c|}{\multirow{4}{*}{C+K}} & SW~\cite{sw} & 41.9 \\
		& \multicolumn{3}{c|}{} & CRDA~\cite{crda} & 43.6 \\
		& \multicolumn{3}{c|}{} & UMT~\cite{umt} & 47.0 \\
		& \multicolumn{3}{c|}{} & UBT~\cite{ubt} (Baseline)& 47.6 \\ \hline
		\multirow{6}{*}{MSDA} & \multicolumn{3}{c|}{\multirow{6}{*}{C+K}} & MDAN~\cite{mdan} & 43.2 \\
		& \multicolumn{3}{c|}{} & M$^3$SDA~\cite{m3sda} & 44.1 \\
		& \multicolumn{3}{c|}{} & DMSN~\cite{dmsn} & 49.2 \\
		& \multicolumn{3}{c|}{} & \textbf{HTRM (Ours)} & 52.9 \\
		& \multicolumn{3}{c|}{} & \textbf{AMSD (Ours)} & 56.8 \\
		& \multicolumn{3}{c|}{} & \textbf{TRKP (Ours)} & \textbf{58.4} \\ \hline
		Oracle & \multicolumn{3}{c|}{BDD100K} & FRCNN~\cite{frcnn} & 60.2 \\ \hline
	\end{tabular}
	\caption{Results on cross camera adaptation. `C' and `K' indicate Cityscapes and KITTI respectively, which constitute source domains. BDD100K is the target domain. AP (\%) of \emph{car} is reported.}
	\label{table:car}
\end{table}

\noindent\textbf{Settings.} 
The images captured by different cameras incur the domain shift problem due to various settings of camera parameters, viewpoints and scenes during data collection. To address this concern, we evaluate our method in the setting of cross camera adaptation. By following~\cite{dmsn}, we select Cityscapes~\cite{cityscapes} and KITTI~\cite{kitti} as the source domains and BDD100K~\cite{bdd} as the target domain, and meanwhile only use the images from the $\emph{car}$ category for training and evaluation. Cityscapes~\cite{cityscapes} is a benchmark for semantic urban scene understanding and KITTI~\cite{kitti} is a widely used dataset for autonomous driving, containing 2,975 and 7,481 annotated training images, respectively. BDD100K is a large-scale dataset for autonomous driving, where only the $\emph{daytime}$ subset is adopted, including 36,728 unlabeled images for training and 5,258 validation images for evaluation. The widely used average precision (AP) is adopted as the evaluation metric.


\noindent\textbf{Results.} As shown in Table~\ref{table:car}, the previous DAOD methods, which simply combine Cityscapes and KITTI (see the row in \emph{``Source Combined''}) during training, generally report worse performance compared to those only adopt Cityscapes (see the row in \emph{``Single Source''}). The reason lies in that knowledge transferred from Cityscapes to BDD100K is probably interfered by the domain shift between Cityscapes and KITTI, resulting in severe knowledge degradation during adaptation. Despite of increasing amount of data in multiple sources, most existing MSDA based methods only achieve minor gains or perform even worse, compared to the source combined approaches. By contrast, our method improves the accuracy by a large margin. For instance, the AP by applying TRKP is 9.2\% higher than the second best, \ie DMSN. It is worth noting that our method is based on the UBT baseline. When separately applying the proposed AMSD and HTRM modules to UBT, the gains are 5.3\% and 9.2\%, respectively, clearly showing their effectiveness. By combining AMSD and HTRM, TRKP achieves an AP of 58.4\%, reaching a new state-of-the-art, which reduces the gap with Oracle (full supervision) to 1.8\%.

\subsection{Cross Time Adaptation} 
\label{sec:exp2}

\noindent\textbf{Settings.} 
In real-world applications, a detector is often deployed at different time, where changes in illumination and scene can be extremely large. To evaluate the performance of our method against such a factor, we follow the setting in \cite{dmsn} to adapt knowledge learned in the daytime and nighttime to corner cases, \ie at dawn or dusk.
Concretely, BDD100K~\cite{bdd} is divided into three subsets by time, including \emph{daytime}, \emph{night}, \emph{dawn/dusk}.
36,728 images in the \emph{daytime} and 27,971 images at \emph{night} constitute two source domains.
Images collected by excluding the ones in the daytime and nighttime are relatively few, where 5,027 unlabeled images are used for training and 778 validation images for evaluation at \emph{dawn/dusk} as the target domain. The mean average precision (mAP) over 10 categories is reported for comparison.


\begin{table}[!t]
	\centering
	\footnotesize
		\begin{tabular}{c|c|c|c}
			\hline
			Setting & Source & Method & mAP \\ \hline
			\multirow{3}{*}{Source Only} & D & \multirow{3}{*}{FRCNN~\cite{frcnn}} &  30.4 \\
			& N &  &  25.0 \\
			& D+N &  &  28.9 \\ \hline
			\multirow{5}{*}{Single Source} & \multirow{5}{*}{D} & SW~\cite{sw} &  31.4 \\
			&  & GPA~\cite{gpa} &  31.8 \\
			&  & CRDA~\cite{crda} &  31.2 \\
			&  & UMT~\cite{umt} &  33.8 \\
			&  & UBT~\cite{ubt} (Baseline) &  33.2 \\ \hline
			\multirow{5}{*}{Single Source} & \multirow{5}{*}{N} & SW~\cite{sw} &  26.9 \\
			&  & GPA~\cite{gpa} &  27.6 \\
			&  & CRDA~\cite{crda} &  28.4 \\
			&  & UMT~\cite{umt} &  21.6 \\
			&  & UBT~\cite{ubt} (Baseline)&  24.2 \\ \hline
			\multirow{5}{*}{Source Combined} & \multirow{5}{*}{D+N} & SW~\cite{sw} &  29.9 \\
			&  & GPA~\cite{gpa} &  30.6 \\
			&  & CRDA~\cite{crda} &  30.2 \\
			&  & UMT~\cite{umt} &  33.5 \\
			&  & UBT~\cite{ubt} (Baseline)&  33.1 \\ \hline
			\multirow{6}{*}{MSDA} & \multirow{6}{*}{D+N} & MDAN~\cite{mdan} &  27.6 \\
			&  & M$^{3}$SDA~\cite{m3sda} &  26.5 \\
			&  & DMSN~\cite{dmsn} &  35.0 \\
			&  & \textbf{HTRM (Ours)} &  35.5 \\
			&  & \textbf{AMSD (Ours)} &  38.0 \\
			&  & \textbf{TRKP (Ours)} & \textbf{39.8} \\ \hline
			Oracle & BDD100K  & FRCNN~\cite{frcnn} &  26.6 \\ \hline
	\end{tabular}
	\caption{Results on cross time adaptation. `D' and `N' indicate the \emph{daytime} and \emph{night} subsets of BDD100K, respectively. mAP (\%) over 10 categories on BDD100K \emph{dawn/dusk} is reported.}
	\label{table:bdd}
\end{table}

\noindent\textbf{Results.}
The results on cross time adaptation are summarized in Table~\ref{table:bdd}, where more detailed comparisons are provided in the \emph{supplementary material} due to space limit.
As shown in Table~\ref{table:bdd}, previous DAOD methods fail to boost the performance when using images from both the \emph{daytime} and \emph{night} subsets, due to the interference of the large discrepancy between the two domains. By multi-source disentanglement, our TRKP improves the performance by  large margins, \eg 4.8\% higher than the second best based on DMSN. The HTRM and AMSD modules also achieve remarkable gains in performance. Specifically, AMSD disentangles multiple sources and prevents the interference among them, thus improving the UBT baseline by 4.9\%. HTRM performs re-weighting at the global level, yielding better performance than DMSN~\cite{dmsn} that adopts the dynamic weighting strategy. Besides, it is worth noting that TRKP exceeds Oracle significantly and boosts the detection accuracy to 39.8\% in mAP. The relatively poor performance of Oracle is owing to insufficient training images in the target domain, and our remarkable performance improvement shows the effectiveness of transfer learning in such situations by target-relevant knowledge adaptation.

\subsection{Extension to Mixed Domain Adaptation}
\label{sec:exp3}
\noindent\textbf{Settings.}
As there always exist more than one factors leading to domain shift in practice, we extend existing settings of cross camera/time adaptions with only two source domains, and present a new setting by considering a more complex case with mixed domain gaps. Specifically, based on the scene adaptation scenario in~\cite{crda} that chooses Cityscapes~\cite{cityscapes} as the source and BDD100K~\cite{bdd} as the target, we employ MS COCO~\cite{coco} and Synscapes~\cite{synscapes} as two extra sources. MS COCO contains common scenes distinct from street views and Synscapes is a synthetic dataset, both of which enlarge the data scale and bring in more kinds of domain gaps and category shifts. 2,975/71,749/25,000 images from Cityscapes/MS COCO/Synscapes are used for training. 36,728 images in the \emph{daytime} subset from BDD100K are used as unlabeled target data.
5,258 images from BDD100K in the \emph{daytime} subset are used for evaluation. mAP over 7 classes is reported.

\begin{table}[]
\footnotesize
\centering
\begin{tabular}{c|c|c|c}
\hline
Setting & Source & Method & mAP \\ \hline
Source Only & C & FRCNN~\cite{frcnn} & 23.4 \\
Single Source & C & UBT~\cite{ubt} (Baseline)& 29.7 \\ \hline
Source Only & C+M & FRCNN~\cite{frcnn} & 29.7 \\
Source Combined & C+M & UBT~\cite{ubt} (Baseline) & 18.5 \\
MSDA & C+M & \textbf{TRKP (Ours)} & \textbf{35.3} \\ \hline
Source Only & C+M+S & FRCNN~\cite{frcnn} & 30.9 \\
Source Combined & C+M+S & UBT~\cite{ubt} (Baseline)& 25.1 \\
MSDA & C+M+S & \textbf{TRKP (ours)} & \textbf{37.1} \\ \hline
Oracle & BDD100K & FRCNN~\cite{frcnn} & 38.6 \\ \hline
\end{tabular}
\caption{Results on mixed domain adaptation. `C'/`M'/`S' indicate Cityscapes/MS COCO/Synscapes, respectively.}
	\label{table:threesource}
\end{table}

\noindent\textbf{Results.}
As summarized in Table~\ref{table:threesource}, by adopting more sources, the performance of the source only detector, \ie FRCNN, is consistently improved. However, the source combined method, \ie UBT, performs poorly due to severe negative transfers caused by mixed domain gaps. In contrast, TRKP achieves a significant performance gain, \eg 5.6\% in mAP when using two sources, and 6.2\% in mAP for three sources, demonstrating its effectiveness when applying to mixed source domains.


\subsection{Ablation Study}
\label{sec:exp4}
We detailedly analyze the modules and hyper-parameters of TRKP in the setting of Cross Time Adaptation.


\noindent\textbf{On Disentanglement.} 
As displayed in Table~\ref{table:disabla}, training a separated detector for each source domain performs much worse than training a common backbone with combined sources, showing the necessity of a shared feature extractor. The multi-head structure also contributes, improving the mAP by 1.3\%. When performing AMSD on the classification head and regression head, mAPs are boosted by 2.1\% and 1.5\% respectively, highlighting the advantage of using adversarial disentanglement. A combination of them further promotes the accuracy.


\noindent\textbf{On Hyper-Parameters.}
As described in Sec.~\ref{sec:disentanglement}, $\mu$ and $\lambda$ control the magnitude of AMSD . As shown in Table~\ref{table:abla}, TRKP achieves the best result when $\mu=0.01$ and $\lambda=0.2$. As for HTRM, we study the effect of the number of neighbors $K'$, where the best result is reached when $K'=5$. Moreover, HTRM focuses on mining source-target relevance at the instance level, rather than at the image-level as in most existing MSDA approaches \cite{distill, nips18}. To validate the impact of instance-level relevance, we report the mAPs by performing HTRM at different levels. As in Table~\ref{table:abla}, HTRM clearly performs better at the instance level, which makes sense as object detection is an instance-aware task.

\begin{table}[!t]
	\centering
	\footnotesize
	\begin{tabular}{cccc|c}
		\hline
		\multicolumn{1}{c}{Shared Feature} & Multi-head & Cls & Reg & mAP \\ \hline
		& $\checkmark$ & $\checkmark$ & $\checkmark$ & 24.7 \\
		$\checkmark$ & &  &  & 33.1 \\
		$\checkmark$ & $\checkmark$ &  &  & 34.4 \\
		$\checkmark$ & $\checkmark$ & $\checkmark$ &  & 36.5 \\
		$\checkmark$ & $\checkmark$ &  & $\checkmark$ & 35.9 \\
		$\checkmark$ & $\checkmark$ & $\checkmark$ & $\checkmark$ & \textbf{38.0} \\ \hline
	\end{tabular}
	\caption{mAP (\%) by performing AMSD on different structures. Shared Feature refers to training with shared backbone and RPN. Multi-head indicates assigning each source an independent RoI head. Cls/Reg refer to applying disentanglement on the classification/regression heads.}
	\label{table:disabla}
\end{table}

\begin{table}[]
	\centering
	\footnotesize
	\begin{tabular}{ccc|ccc}
		\hline
		\multicolumn{3}{c|}{AMSD} & \multicolumn{3}{c}{HTRM} \\ \hline
		$\mu$ & \multicolumn{1}{c|}{$\lambda$} & mAP & Features & \multicolumn{1}{c|}{$K'$} & mAP \\ \hline
		0.05 & \multicolumn{1}{c|}{0.2} & 36.8 & Image-level & \multicolumn{1}{c|}{5} & 32.6 \\
		0.002 & \multicolumn{1}{c|}{0.2} & 37.0 & Instance-level & \multicolumn{1}{c|}{3} & 34.9 \\
		0.01 & \multicolumn{1}{c|}{0.2} & \textbf{38.0} & Instance-level & \multicolumn{1}{c|}{5} & \textbf{35.5} \\
		0.01 & \multicolumn{1}{c|}{1.0} & 37.2 & Instance-level & \multicolumn{1}{c|}{10} & 35.2 \\
		0.01 & \multicolumn{1}{c|}{0.04} & 36.7 & Instance-level & \multicolumn{1}{c|}{30} & 34.6 \\ \hline
	\end{tabular}
	\caption{mAP (\%) of ablation studies on AMSD and HTRM.}
	\label{table:abla}
\end{table}




\section{Conclusion}
\label{sec:conclusion}
In this paper, we present a novel multi-source domain adaptation approach for object detection. To avoid knowledge degradation, we propose an adversarial multi-source disentanglement module and a holistic target-relevant mining scheme to preserve target-relevant knowledge during adaption.
Extensive experiments clearly show the effectiveness of our method compared to the state-of-the-art. Besides, we apply our method to a harder scenario with mixed sources and provide a competitive baseline.

\section*{Acknowledgement}
This work is partly supported by the National Natural Science Foundation of China (No. 62022011, No. 61876176 and U1813218), the Guangdong NSF Project (2020B1515120085), the Shanghai Committee of Science and Technology, China (21DZ1100100), the Shenzhen Research Program (RCJC20200714114557087), the Research Program of State Key Laboratory of Software Development Environment (SKLSDE-2021ZX-04), the Fundamental Research Funds for the Central Universities, and the Joint Lab of CASHK.

{\small
\bibliographystyle{ieee_fullname}
\bibliography{paper}
}

\clearpage
\appendix
\setcounter{section}{0}
\setcounter{table}{0}
\setcounter{figure}{0}
\renewcommand{\thesection}{\Alph{section}}
\renewcommand{\thefigure}{\Alph{figure}}
\renewcommand{\thetable}{\Alph{table}}

\noindent{\large\textbf{Appendix}}

\begin{figure*}[!h]
	\centering 
	\includegraphics[width=0.8\linewidth]{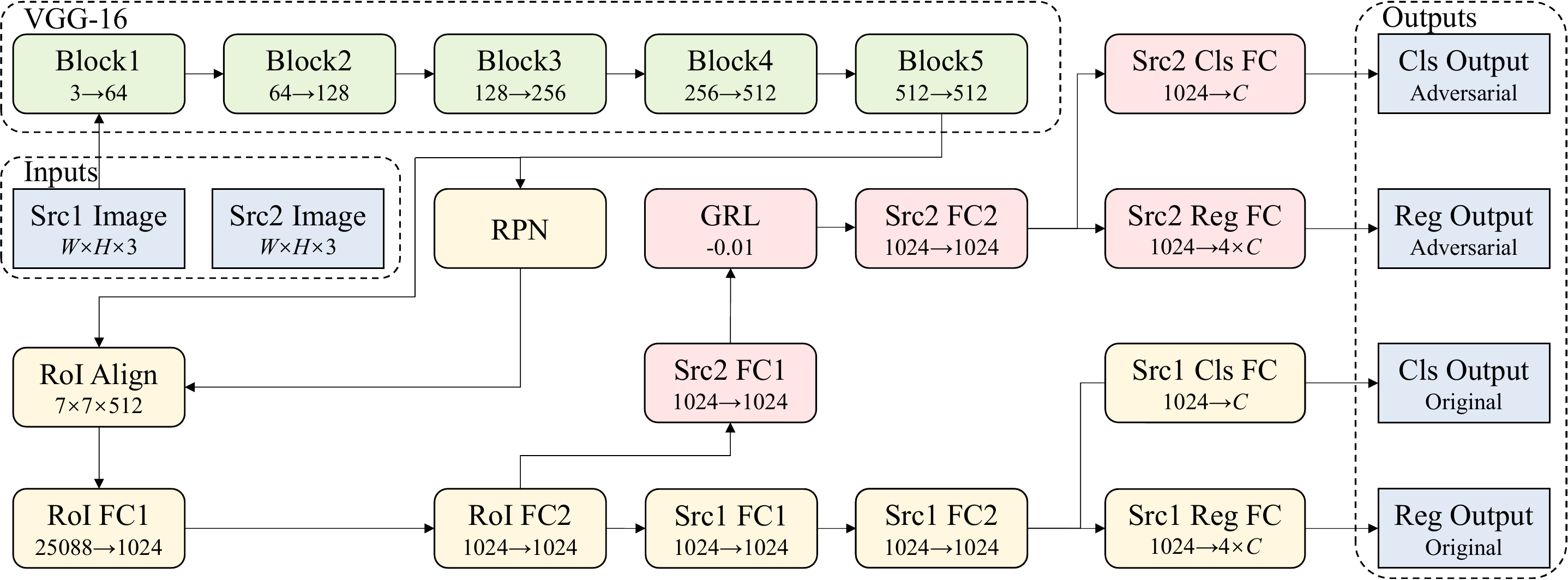}
	\caption{Illustration of the detailed network architecture of the teacher detector TeDet$(\cdot)$ with the AMSD module for two source domains. The configuration and the sizes of channels/feature maps are also presented. ``Block'' stands for the convolutional network layers of VGG~\cite{vgg} and ``FC'' refers to the fully-connected layer. ``$W\times{H}$'' and ``$C$'' indicate the image size and the number of object categories, respectively. }
	\label{fig:network}
\end{figure*}

In this supplementary material, we provide more implementation details of the detector in \cref{sec:1}, detailed experimental results for the settings of Cross Time Adaptation and Mixed Domain Adaptation in \cref{sec:2}, visualization results of the HTRM module in \cref{sec:vis_htrm} as well as discussion on limitations of our approach in \cref{sec:dis}.

\section{More Implementation Details}
\label{sec:1}

In this section, we provide more implementation details about the network structure of the teacher detector TeDet$(\cdot)$. Since the student detector StDet$(\cdot)$ shares the same structure as the teacher detector, we therefore only describe the details of TeDet$(\cdot)$. Without loss of generality, we consider TeDet$(\cdot)$ with the AMSD module for two source domains. As shown in Fig.~\ref{fig:network}, TeDet$(\cdot)$ consists of the VGG-16 backbone, RPN, RoI Align, RoI feature extractor, \emph{GRL} and the multiple heads, where their configurations and the sizes of channels/feature maps are also displayed.

Images from each source domain are applied to train the corresponding head and perform adversarial learning on the other heads.
Given an image from the target domain, the multiple heads make predictions simultaneously based on proposals from the shared RPN.
On each proposal, the predicted classification and regression results from multi-heads are aggregated by averaging before non-maximum suppression. We implement the overall training process of the teacher-student framework based on the open source\footnote{https://github.com/facebookresearch/unbiased-teacher} of UBT~\cite{ubt}.
In all experiments, we adopt VGG-16~\cite{vgg} pretrained on ImageNet~\cite{imagenet} as the backbone. 

\section{Detailed Experimental Results}
\label{sec:2}
In this section, we display more experimental results for the settings of Cross Time Adaptation in \cref{subsec:cta} and Extension to Mixed Domain Adaptation in \cref{subsec:mda}, respectively. 

\subsection{Cross Time Adaptation}
\label{subsec:cta}

As demonstrated in Table~\ref{table:bdddetail}, we report the AP of all categories on the BDD100K \emph{dawn/dusk} subset.
By following \cite{dmsn}, the result on the category 'train' is not reported.
The proposed TRKP approach outperforms the other counterparts for most categories.
Both AMSD and HTRM improve the detection performance for almost all the categories and achieve the best result when they are combined.
%

\begin{table*}[!thp]
	\centering
	\footnotesize
	\setlength{\tabcolsep}{2mm}{
		\begin{tabular}{c|c|c|cccccccccc|c}
			\hline
			Setting & Source & Method & bike & bus & car & motor & person & rider & light & sign & train & truck & mAP \\ \hline
			\multirow{3}{*}{\begin{tabular}[c]{@{}c@{}}Source\\ Only\end{tabular}} & D & \multirow{3}{*}{FRCNN~\cite{frcnn}} & 35.1 & 51.7 & 52.6 & 9.9 & 31.9 & 17.8 & 21.6 & 36.3 & - & 47.1 & 30.4 \\
			& N &  & 27.9 & 32.5 & 49.4 & 15.0 & 28.7 & 21.8 & 14.0 & 30.5 & - & 30.7 & 25.0 \\
			& D+N &  & 31.5 & 46.9 & 52.9 & 8.4 & 29.5 & 21.6 & 21.7 & 34.3 & - & 42.2 & 28.9 \\ \hline
			\multirow{6}{*}{\begin{tabular}[c]{@{}c@{}}Single\\ Source\end{tabular}} & \multirow{6}{*}{D} & SW~\cite{sw} & 34.9 & 51.2 & 52.7 & 15.1 & 32.8 & 23.6 & 21.6 & 35.6 & - & 47.1 & 31.4 \\
			&  & SCL~\cite{scl} & 29.1 & 51.3 & 52.8 & 17.2 & 32.0 & 19.1 & 21.8 & 36.3 & - & 47.2 & 30.7 \\
			&  & GPA~\cite{gpa} & 36.6 & 52.1 & 53.1 & 15.6 & 33.0 & 23.0 & 21.7 & 35.4 & - & 48.0 & 31.8 \\
			&  & CRDA~\cite{crda} & 32.8 & 51.4 & 53.0 & 15.4 & 32.5 & 22.3 & 21.2 & 35.4 & - & 47.9 & 31.2 \\
			&  & UMT~\cite{umt} & 39.7 & 52.3 & 56.1 & 14.2 & 35.7 & 23.7 & 31.5 & 42.2 & - & 42.4 & 33.8 \\
			&  & UBT~\cite{ubt} (Baseline) & 37.4 & 52.3 & 56.6 & 14.3 & 35.0 & 22.9 & 31.1 & 40.3 & - & 42.6 & 33.2 \\ \hline
			\multirow{6}{*}{\begin{tabular}[c]{@{}c@{}}Single\\ Source\end{tabular}} & \multirow{6}{*}{N} & SW~\cite{sw} & 31.4 & 38.2 & 51.0 & 9.9 & 29.5 & 22.2 & 18.7 & 32.5 & - & 35.7 & 26.9 \\
			&  & SCL~\cite{scl} & 25.3 & 31.7 & 49.3 & 8.9 & 25.8 & 21.2 & 15.0 & 28.6 & - & 26.2 & 23.2 \\
			&  & GPA~\cite{gpa} & 32.7 & 38.3 & 51.8 & 14.1 & 29.0 & 21.5 & 17.1 & 31.1 & - & 40.0 & 27.6 \\
			&  & CRDA~\cite{crda} & 32.3 & 45.1 & 51.6 & 7.2 & 29.2 & 24.9 & 19.9 & 33.0 & - & 41.1 & 28.4 \\
			&  & UMT~\cite{umt} & 37.9 & 18.4 & 50.4 & 8.8 & 24.7 & 11.6 & 15.1 & 30.1 & - & 19.4 & 21.6 \\
			&  & UBT~\cite{ubt} (Baseline) & 42.7 & 18.8 & 52.5 & 8.2 & 26.5 & 20.0 & 19.7 & 29.5 & - & 23.7 & 24.2 \\ \hline
			\multirow{6}{*}{\begin{tabular}[c]{@{}c@{}}Source\\ Combined\end{tabular}} & \multirow{6}{*}{D+N} & SW~\cite{sw} & 29.7 & 50.0 & 52.9 & 11.0 & 31.4 & 21.1 & 23.3 & 35.1 & - & 44.9 & 29.9 \\
			&  & SCL~\cite{scl} & 33.9 & 47.8 & 52.5 & 14.0 & 31.4 & 23.8 & 22.3 & 35.4 & - & 45.1 & 30.9 \\
			&  & GPA~\cite{gpa} & 31.7 & 48.8 & 53.9 & 20.8 & 32.0 & 21.6 & 20.5 & 33.7 & - & 43.1 & 30.6 \\
			&  & CRDA~\cite{crda} & 25.3 & 51.3 & 52.1 & 17.0 & 33.4 & 18.9 & 20.7 & 34.8 & - & 47.9 & 30.2 \\
			&  & UMT~\cite{umt} & 42.3 & 48.1 & 56.4 & 13.5 & 35.3 & 26.9 & 31.1 & 41.7 & - & 40.1 & 33.5 \\
			&  & UBT~\cite{ubt} (Baseline) & 40.5 & 49.9 & 56.4 & 14.5 & 33.7 & 23.6 & 30.4 & 40.0 & - & 41.6 & 33.1 \\ \hline
			\multirow{6}{*}{MSDA} & \multirow{6}{*}{D+N} & MDAN~\cite{mdan} & 37.1 & 29.9 & 52.8 & 15.8 & 35.1 & 21.6 & 24.7 & 38.8 & - & 20.1 & 27.6 \\
			&  & M$^{3}$SDA~\cite{m3sda} & 36.9 & 25.9 & 51.9 & 15.1 & 35.7 & 20.5 & 24.7 & 38.1 & - & 15.9 & 26.5 \\
			&  & DMSN~\cite{dmsn} & 36.5 & 54.3 & 55.5 & 20.4 & 36.9 & \textbf{27.7} & 26.4 & 41.6 & - & 50.8 & 35.0 \\
			&  & \textbf{HTRM (Ours)} & 41.6 & 50.9 & 58.3 & 21.5 & 37.6 & 24.7 & 35.3 & 43.6 & - & 41.3 & 35.5 \\
			&  & \textbf{AMSD (Ours)} & 44.0 & 55.3 & 60.1 & 17.7 & 39.8 & 26.7 & 37.9 & 46.9 & - & 51.2 & 38.0 \\
			&  & \textbf{TRKP (Ours)} & \textbf{48.4} & \textbf{56.3} & \textbf{61.4} & \textbf{22.5} & \textbf{41.5} & 27.0 & \textbf{41.1} & \textbf{47.9} & - & \textbf{51.9} & \textbf{39.8} \\ \hline
			Oracle & BDD100K & FRCNN~\cite{frcnn} & 27.2 & 39.6 & 51.9 & 12.7 & 29.0 & 15.2 & 20.0 & 33.1 & - & 37.5 & 26.6 \\ \hline
	\end{tabular}}
	\caption{Detailed results for the setting of Cross Time Adaptation. `D' and `N' indicate the \emph{daytime} and \emph{night} subsets of BDD100K. mAP (\%) for all the classes and detailed AP (\%) of each individual category on BDD100K \emph{dawn/dusk} are reported. Best in bold.}
\label{table:bdddetail}
\end{table*}

\subsection{Extension to Mixed Domain Adaptation}
\label{subsec:mda}

More results for the setting of Mixed Domain Adaptation are summarized in Table~\ref{table:threesourcedetail}, where
the category ``train" with very few instances is ignored as in~\cite{crda}.
With more available sources, the detection performance is consistently improved for most categories except for ``rider'', where the results drop when introducing more data. The reason behind probably lies in the huge domain gap and category shift between the source domains w.r.t. the `rider' class. Despite of that, our method reaches the best results in most cases, showing its effectiveness. 

\begin{table*}[!thp]
	\footnotesize
	\centering
	\begin{tabular}{c|c|c|cccccccc|c}
		\hline
		Setting & Source & Method & person & car & train & rider & truck & motor & bicycle & bus & mAP \\ \hline
		Source Only & C & FRCNN~\cite{frcnn} & 26.9 & 44.7 & - & 22.1 & 17.4 & 17.1 & 18.8 & 16.7 & 23.4 \\
		Single Source & C & UBT~\cite{ubt} (Baseline)& 37.8 & 50.9& - & \textbf{38.2} & 21.3 & 19.9 & 29.9 & 10.9& 29.7 \\ \hline
		Source Only & C+M & FRCNN~\cite{frcnn} & 35.2 & 49.5 & - & 26.1 & 25.8 & 18.9 & 26.1 & 26.5  & 29.7 \\
		Source Combined & C+M & UBT~\cite{ubt} (Baseline) & 30.7 & 28.0 & - & 3.9 & 11.2 & 19.2 & 17.8 & 18.7 & 18.5 \\
		MSDA & C+M & \textbf{HTRM (Ours)} & 34.6 & 48.3 & - & 20.2 & 21.7 & 26.7 & 32.0 & 34.1 & 31.1 \\
		MSDA & C+M & \textbf{AMSD (Ours)} & 38.6 & 52.1 & - & 28.2 & 22.9 & 24.9 & 28.5 & 33.3 & 32.6 \\
		MSDA & C+M & \textbf{TRKP (Ours)} & 39.2 & 53.2 & - & 32.4 & 28.7 & 25.5 & 31.1 & 37.4 & 35.3 \\ \hline
		Source Only & C+M+S & FRCNN~\cite{frcnn} & 36.6 & 49.0 & - & 22.8 & 24.9 & 26.9 & 28.4 & 27.7 & 30.9 \\
		Source Combined & C+M+S & UBT~\cite{ubt} (Baseline)& 32.7 & 39.6 & - & 6.6 & 21.2 & 21.3 & 25.7 & 28.5 & 25.1 \\
		MSDA & C+M+S & \textbf{HTRM (ours)} & 37.7 & 50.2 & - & 20.5 & \textbf{32.7} & 27.0 & 30.4 & 35.7 & 33.5 \\
		MSDA & C+M+S & \textbf{AMSD (ours)} & 40.1 & 52.8 & - & 25.3 & 25.9 & 29.1 & 31.8 & 36.2 & 34.5 \\
		MSDA & C+M+S & \textbf{TRKP (ours)} & \textbf{40.2} & \textbf{53.9} & - & 31.0 & 30.8 & \textbf{30.4} & \textbf{34.0} & \textbf{39.3} &\textbf{37.1} \\ \hline
		Oracle & BDD100K & FRCNN~\cite{frcnn} & 35.3 & 53.9 & - & 33.2 & 46.3 & 25.6 & 29.3 & 46.7 & 38.6 \\ \hline
	\end{tabular}
	\caption{Detailed results for the setting of Mixed Domain Adaptation. `C'/`M'/`S' indicate Cityscapes/MS COCO/Synscapes, respectively. mAP (\%) and detailed AP (\%) of each category on BDD100K \emph{daytime} are reported.}
	\label{table:threesourcedetail}
\end{table*}




\begin{figure*}[!t]
	\centering
	\includegraphics[width=0.75\linewidth]{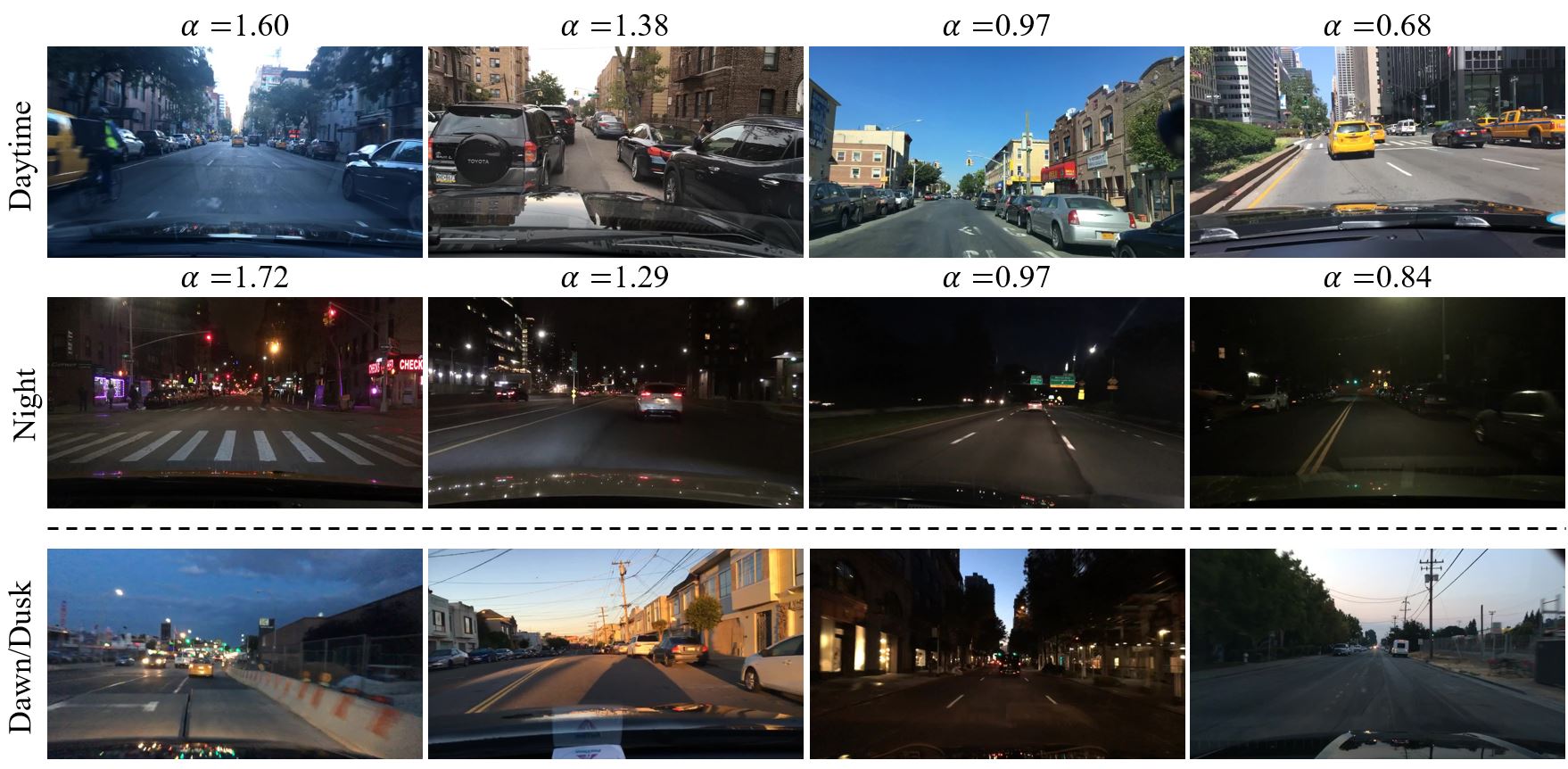}
	\caption{Visualization of the source images ranked by weights generated via HTRM on BDD100K. With a larger target-relevance weight $\alpha$, the corresponding source image appears more similar to the images from the target domain, \emph{i.e.} \emph{dawn/dusk}.}
	\label{fig:vis}
\end{figure*}

\section{Visualization of HTRM}
\label{sec:vis_htrm}

To display the effectiveness of the HTRM module, we demonstrate the images with different target-relevance weights in the Cross Time Adaptation setting on the BDD100K dataset. 

Recall that the source domains consist of images from \emph{Daytime} and \emph{Night}, and the target domain from \emph{Dawn/Dust}. As shown in Fig.~\ref{fig:vis}, the source image with a larger weight $\alpha$ clearly has a more similar appearance to those from the target, in regard of the illumination condition.

\section{Discussion on limitations.}
\label{sec:dis}
The existing study~\cite{dmsn} considers two sources (cross camera and cross time). Although we extend it to a harder case with three sources, the experimental setting of multi-source DAOD is still at street views. We will consider more source domains and larger domain gaps to further improve the generality in our future work.

\end{document}